\documentclass[11pt]{article}
\usepackage[margin=1in]{geometry}
\usepackage{relsize}
\usepackage{amssymb}
\usepackage{amsmath}
\usepackage{enumitem}
\usepackage{graphicx}
\usepackage[linktocpage]{hyperref}
\usepackage{amsthm}
\usepackage{comment}
\usepackage{subfig}
\usepackage{wrapfig}
\usepackage{todonotes}
\usepackage[maxbibnames=99, backend=biber,style=alphabetic]{biblatex}
\bibliography{references} 
\begin{document}

\title{Improving Radiography Machine Learning Workflows via Metadata Management for Training Data Selection}

\author{Mirabel Reid\\mreid48@gatech.edu\\ Georgia Tech, LANL\thanks{Work completed while at Los Alamos National Laboratory}\and 
Christine Sweeney \\ cahrens@lanl.gov \\ LANL\and
Oleg Korobkin \\ korobkin@lanl.gov\\LANL
}

\maketitle
\begin{abstract}
Most machine learning models require many iterations of hyper-parameter tuning, feature engineering, and debugging to produce effective results. As machine learning models become more complicated, this pipeline becomes more difficult to manage effectively. In the physical sciences, there is an ever-increasing pool of metadata that is generated by the scientific research cycle. Tracking this metadata can reduce redundant work, improve reproducibility, and aid in the feature and training dataset engineering process. In this case study, we present a tool for machine learning metadata management in dynamic radiography. We evaluate the efficacy of this tool against the initial research workflow and discuss extensions to general machine learning pipelines in the physical sciences.\footnote{LA-UR-22-30449}
\end{abstract}

\section{Introduction}
Machine learning (ML) is an increasingly integral part of scientific research. In the natural sciences, ML can recognize patterns in observational data and inform the development of scientific theories~\cite{roscher2020explainable}. In applications such as materials science, an effective machine learning model can dramatically reduce the need for costly experiments and long development cycles~\cite{wei2019machine,carleo2019machine}. It is undeniable that ML can be an effective method for interpreting large volumes of observed or simulated data.

However, the introduction of a complex method like ML comes with challenges. Outside of simple models, most ML models require many iterations of hyper-parameter tuning, feature engineering, and debugging to produce effective results. Because of this, the workflow is often packaged into a pipeline: an automated or semi-automated program which prepares the data, trains, and evaluates the model in one swoop. As machine learning models become more complicated, the pipeline becomes more difficult to manage effectively. The creation of tools which automatically track metadata is an active area of development. Several industry leaders in ML such as Google~\cite{google} and Netflix~\cite{netflix} have built and released their own open source metadata stores.

\begin{figure}[h]
\centering
    \includegraphics[width=0.8\textwidth]{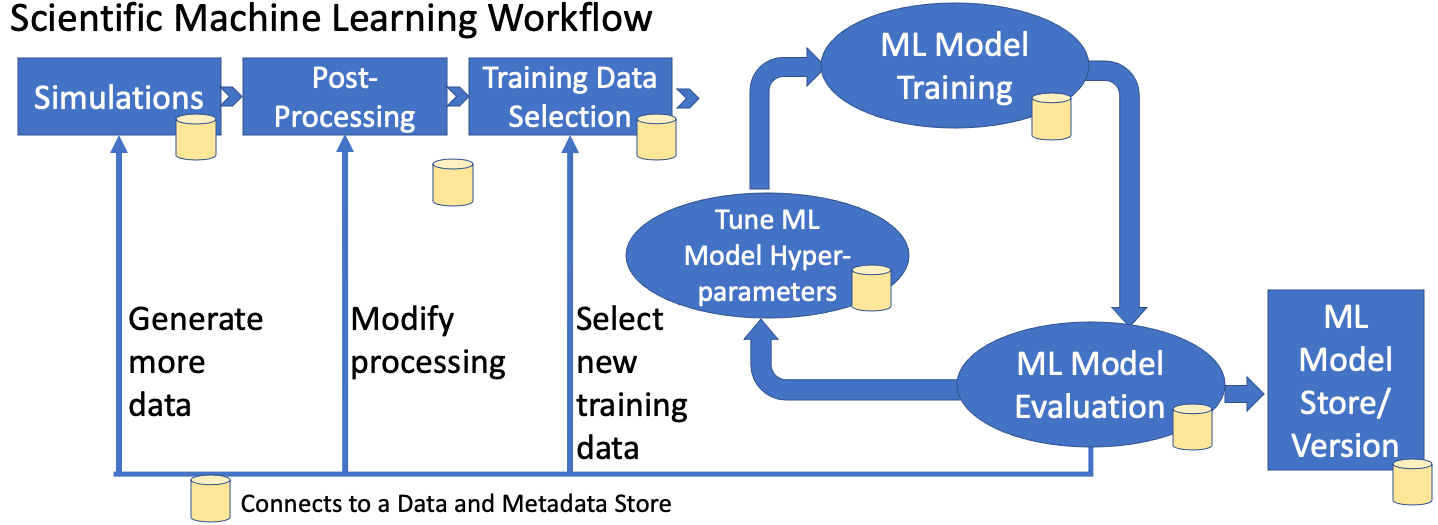}
    \caption{A flowchart describing a generic pipeline for learning on simulation data. As the objectives change over time, each step may be repeated and fine-tuned. Ovals indicate the steps of machine learning training. Each step is marked with a cylinder to indicate that metadata generated from that step is tracked in an external store.}
    \label{fig-ml-pipeline}
    %\alt{A flow chart. The first three boxes are `Simulations', `Post-Processing', and `Training Data Selection'. This feeds into a cycle with three boxes, labeled `Training', `Evaluation', and `Hyperparameter tuning'. The `Evaluation' box has an arrow to the right leading to `ML Model Store and Version'. It also has arrows back to the first three boxes.}
\end{figure}

Most existing tools to manage the ML pipeline are aimed at commercial applications and web-based learning. There is a dearth of metadata management tools for the physical sciences. Figure \ref{fig-ml-pipeline} shows a generic pipeline which scientists may employ in the research cycle. While the scientific research cycle has similarities to the commercial development cycle, there are significant differences that often make commercial tools impractical to apply. For example, these tools tend to emphasize the capability to package and deploy models to production, a step which is often unnecessary in research in the physical sciences~\cite{zaharia2018accelerating}. Additionally, commercial ML applications generally aim to improve model performance in order to maximize profits or improve the service~\cite{zaharia2018accelerating}, while scientific applications may also require that the model conforms to natural laws. In many cases, physical scientists use ML models as tools to gain scientific understanding or develop theories about causal relationships~\cite{roscher2020explainable}. This leads to a different development cycle to which a pipeline management tool focused on monitoring performance may not adapt well. 
In addition, existing studies on pipeline management tools, such as ~\cite{hohman2020understanding,amershi2019software}, are aimed at ML developers who are likely already accustomed to using automated ML pipelines. ML management principles that are standard
among ML developers can also apply to scientific workflows.  Metadata tracking helps define what is useful metadata for science

\paragraph{Contribution.} We engineered a tool to store and visualize the metadata generated by a scientific research project with complex metadata management needs. The project studies machine learning in dynamic radiography, an imaging tool used to reconstruct properties of materials undergoing strong deformations. The computer scientists and domain scientists worked together to identify essential metadata tracking capabilities that their previous workflow could not support. These desired capabilities included 1) interactive investigation of the feature space to locate degeneracies and multi-modal regions of the parameter space; 2) the ability to select training data based on visual queries on the feature space; and 3) centralized tracking of training datasets alongside the query parameters used to generate them. We created a tool which simplifies data exploration and training data selection, stores the metadata associated with this process, and can recall this metadata to reproduce previous selections. 

\paragraph{Benefits of Metadata Tracking.} The word `metadata' encompasses a wide variety of statistics, parameters, and artifacts generated during the model training and production process. In this case study, we primarily focus on tracking training dataset generation. It is increasingly common for scientific training data to come wholly or in part from numerical simulations; the data generation step may be iterated many times, and the simulation parameters are additional metadata associated with the data. Once the data is generated, a sub-selection of the data is processed and used to train the model. In the sciences, this step of the pipeline often involves many iterations of qualitative decisions, which presents a unique challenge from a metadata tracking perspective. Nonetheless, there are several vital benefits to a robust metadata tracking system, which we outline below. 

The first is efficiency and avoiding waste. Modern machine learning models take a long time to train and use up a lot of compute resources. Therefore, it is important to keep track of what work has been done so that training is not repeated unnecessarily. An efficient tracking pipeline can also reduce the amount of time spent engineering features. 

The second is reproducibility. Machine learning models can often be obtuse and difficult to interpret. In scientific applications, the pre-processing and training data selection are essential parts of the pipeline, and improvements in this step can be a key contribution in scientific research. Hence, in order to reproduce results, it is important to track all choices made. 

The third benefit we will consider is scientific exploration. In complicated projects where there are many interdependencies, it can be difficult to engineer features and create a training dataset that accurately represents the underlying phenomenon. Tracking, storing, and visualizing a wide variety of metadata can help the scientist recognize holes where additional experiments may be needed. 

\paragraph{Related Work.} Metadata tracking solutions are common in commercial production environments. Several industry leaders in machine learning such as Netflix~\cite{netflix} and Google~\cite{google} have built and released their own pipeline management tools.  Additionally, there are numerous studies on the use of visual analytics to support data preparation~\cite{hohman2020understanding, liu2018steering} and model interpretation~\cite{hohman2018visual} for machine learning. However, most tools studied in these surveys are geared toward commercial ML or ML for data science. Also, these frameworks tend to be as general as possible, while our tool is specialized to the needs of a particular research project.

Our tool expands on tools for visualizing multi-dimensional scientific data such as Cinema~\cite{ahrens2014image} and CrossVis~\cite{steed2020crossvis}, which demonstrate the effectiveness of incorporating interactive parallel coordinates plots for data exploration. 

\section{Methods} \label{sec:methods}
In this case study, we investigate machine learning metadata management in dynamic radiography, create a tool to help manage the ML pipeline and evaluate the efficacy of this tool against the initial research workflow. First, in Section~\ref{sec:usecase} we present the problem and discuss the challenges related to metadata tracking. In Section~\ref{sec:tooldev} we discuss how we worked with the scientists to develop the tool.  In Section~\ref{sec:gui} we describe the features in the graphical user interface (GUI) of the resulting visualization and selection tool.  In Appendix ~\ref{sec:db}, we present the structure of the database back end that manages the metadata. 

\subsection{Use Case} \label{sec:usecase}

\begin{figure}[h]
\centering
\includegraphics[width=0.6\textwidth]{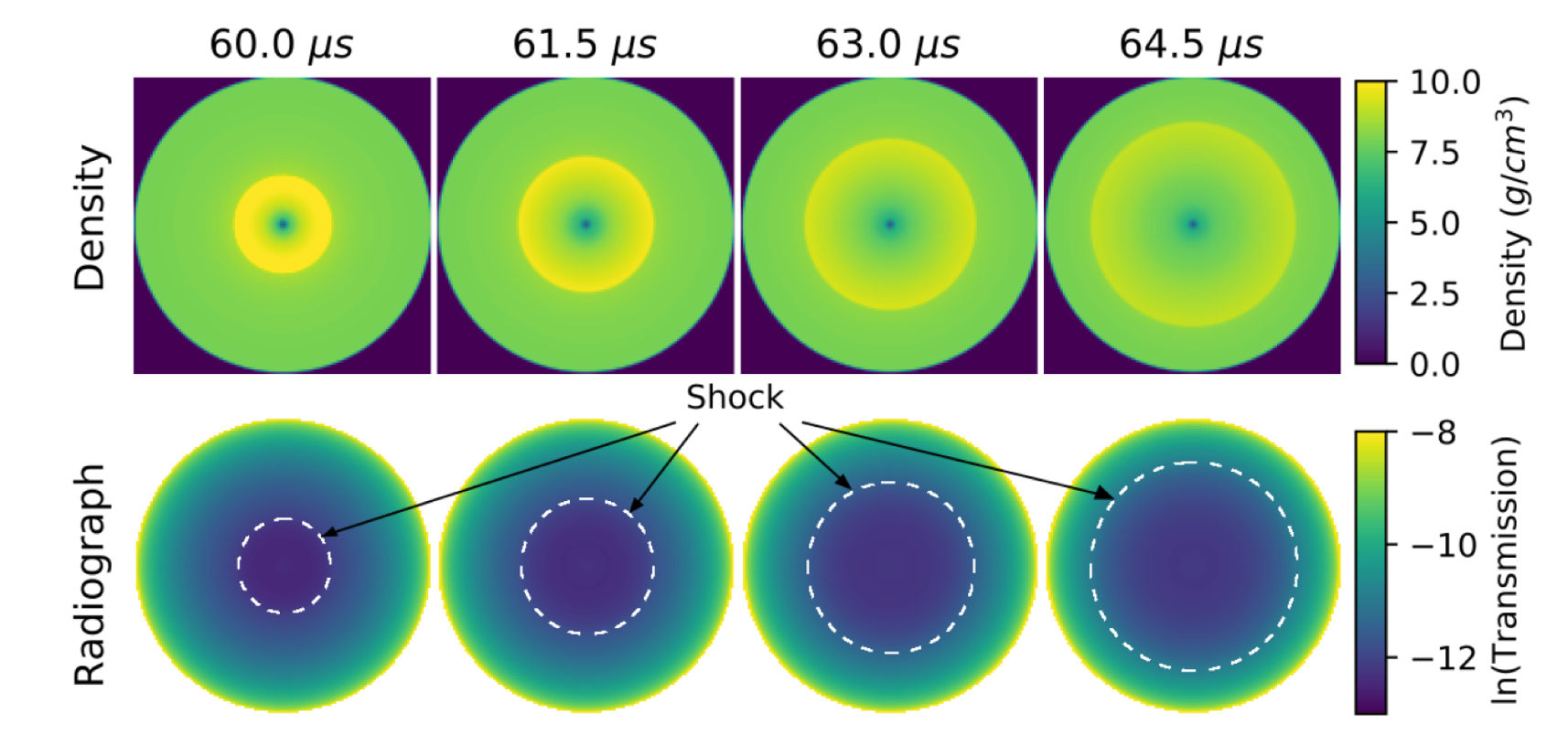}
\caption{A simulated shell implosion at four different time steps. The bottom image shows the original radiograph, and the top image shows the extracted density. The shock, which expands outward over time, is labeled on each radiograph with a dotted line. Image courtesy of~\cite{hossain2022high}.}
\label{fig-radiograph}
%\alt{A 2 by 4 grid of circles. The columns are labeled 60 microseconds, 61.5 microseconds, 63 microseconds, and 64.5 microseconds. The top row is labeled 'density', and the bottom row is labeled 'radiograph'. From left to right, each circle has a smaller circle, labelled 'shock' which expands over time. }
\end{figure}
Radiography is an imaging technique where high-powered X-ray machines are used to probe properties of materials. This technique is used in materials science and shock physics to produce precise estimates of the density field within the object of interest. However, the accuracy of these images is inherently limited by noise; one of the major sources of noise is scattered radiation, which is too complex to model and correct~\cite{hossain2022high}. 

To combat the complexity of recovering the density field from a noisy radiograph, physicists use a database of numerical simulations, each generated by slightly modifying the initial conditions of the material and the physics model. They can then adapt this database to train a ML model to reconstruct density fields. One method to overcome the problem of noise is to train the model using discontinuous features of the simulated radiograph; these features are largely independent of noise, so they are useful for the reconstruction problem. Some examples include the location of the expanding shock in the material and the edge of the shell. 

The use case attempts to recover the density fields from simulated radiographs of shell implosion experiments~\cite{hossain2022high}. The data consist of over 10k two-dimensional shell-implosion simulations; they are parameterized by seven initial conditions which fall into a limited number of categorical values. From these files, the density and the locations of the shock and edge at each of the forty time steps are extracted. The time sequence of shock and edge locations are then provided as features to a Generative Adversarial Network (GAN). 

A challenge in choosing training datasets based on shock and edge features is that the features are not fully descriptive of the density field; that is, two simulations that have similar shock and edge locations at a given time step can have large differences in the density space. Local training data selection can overcome this issue; the model is trained on a dataset consisting of simulations which have large differences in density but small differences in shock and edge. To explore these differences, each simulation in the database is compared against a fixed `Ground Truth' simulation at a fixed time step. This produces a single float for each of the shock difference (`$\delta$ shock'), edge difference (`$\delta$ edge'), and density difference (`$\delta \rho$') at each time step. Such comparisons allow the scientists to make statements about the possibility to recover the density fields from features, or to recover specific parameters based on features alone.
 
\subsection{Tool Development} \label{sec:tooldev}
 We developed a tool to aid in local training data selection for the radiograph use case.  Our tool was designed by an interdisciplinary team, which included computer scientists and domain scientists.  The computer scientists worked with the domain scientist users through meetings and a number of tool demonstrations to obtain requirements for a graphical tool for data exploration and training data selection. The domain scientist users for this tool include a co-author on this paper and another scientist, who are both physicists working in the field of hydrodynamics simulations and radiography.  

As a first step, the computer scientists adapted the data and displayed it using the Cinema Explorer visualization tool~\cite{ahrens2014image}. The benefit of this tool is that it aggregates the metadata from each of the simulations and displays them using a Parallel Coordinates graph, which can visualize broad trends and relationships between metadata parameters. After presenting this tool to the users, the users described the interactive capabilities that would be needed for their application. 

The users noted that the Cinema tool was insufficient to explore the relationship between the shock difference, edge difference, and density difference parameters. While the Parallel Coordinates plot would display these values, a scatter plot would give greater ability to examine the structural relationship between the three parameters. Additionally, without further exploration they were not sure whether the shock difference or edge difference would have greater impact, so they requested that the tool would allow them to re-weight the shock and edge on the fly. Lastly, they mentioned that in order to select points for training data sets, it would be useful to be able to select and save points directly from the plot in the GUI. 

The computer scientists developed a draft iteration of the visualization tool according to this feedback. After loading the data and manipulating it as they would in an experimental setting, the users were able to narrow down additional features which would improve the tool. They noted that the data has an additional dimension - time - which was not accounted for in the initial draft of the tool. It would be useful for them to be able to switch between data taken at different times in the shell implosion. 

In this meeting, the users also explained the process of developing the post-processed data. Ideally, the data is computed against a fixed ground truth time step using a particular distance function (see Appendix ~\ref{sec-pipeline} for the details). However, in order to gain more insight into these relationships, they wanted to try different ground truth simulations and different methods of computing the distance. Hence, the tool would have to house all these different computation methods and visualize them adaptively.  

As a long-term goal of the team, the larger machine learning workflow also needs to be supported.  This includes iterating through various cycles as shown in Figure \ref{fig-ml-pipeline}.  In order to do that, metadata from each step needs to be collected and used.  That includes metadata from using the training data selection tool, so that it is easy and efficient to re-enter the workflow and make modifications.  Thus, additional capability to save the tool GUI settings was added as a requirement.  This capability allows the user to reload the previous training data selections into the tool, modify, then save again.

Finally, to aid training dataset selection from very large sets of simulations, the users requested a capability to do probabilistic subsampling of the selected training dataset in order to keep the set to a desired size.  We present the final visualization tool in the next section.

\subsection{Graphical User Interface} \label{sec:gui}
\begin{figure}[h!]
\centering
    \includegraphics[width=0.85\textwidth]{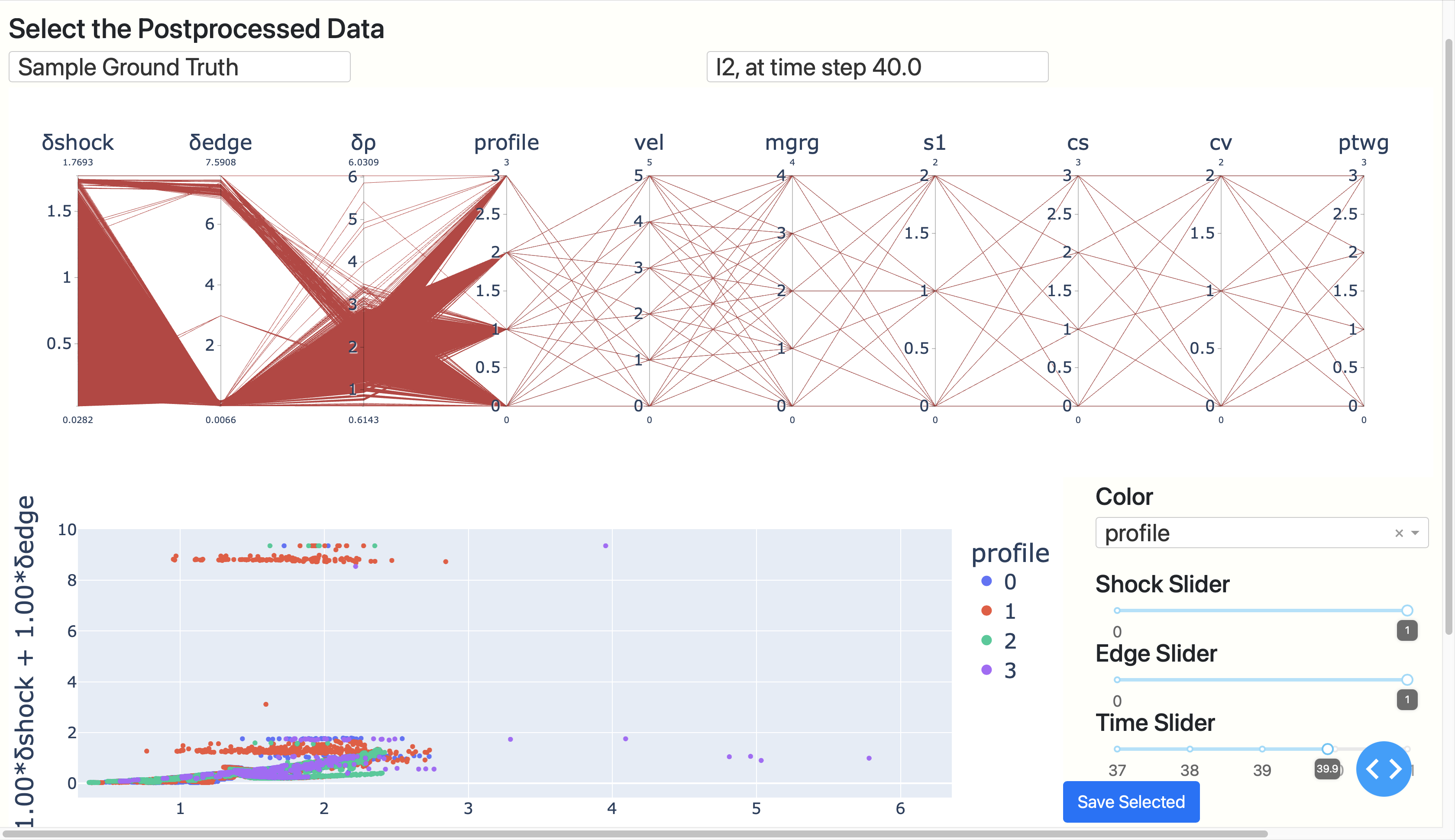}
    \caption{The initial page of the training data selection tool GUI.  At the top, the "Sample Ground Truth" dataset is selected, and the user is viewing the l2 norm between that ground truth dataset and the other available simulation data at time 40.  All available simulation shock, edge and density differences as well as simulation parameter metadata are available for selection in the parallel coordinates plot at the top. Here all values are selected. The scatter plot below allows the user to make the comparison with adjustments of show and edge contributions. }
    %\alt{A screenshot of the initial page of the visualization tool. The title says `Select the Postprocessed Data'. It shows a stacked parallel coordinates plot at the top and a scatter plot below. Above the plots, there are two dropdown menus with selected values `Sample Ground Truth' and `l2, at time step 40.0'. Next to the scatter plot, there is a dropdown menu labelled 'Color' and three sliders labelled `Shock Slider', `Edge Slider' and `Color Slider'. There is a button labelled `Save Selected'.}
    \label{fig-selector-screen}
\end{figure}
\begin{figure}[h]
\centering
    \includegraphics[width=0.85\textwidth]{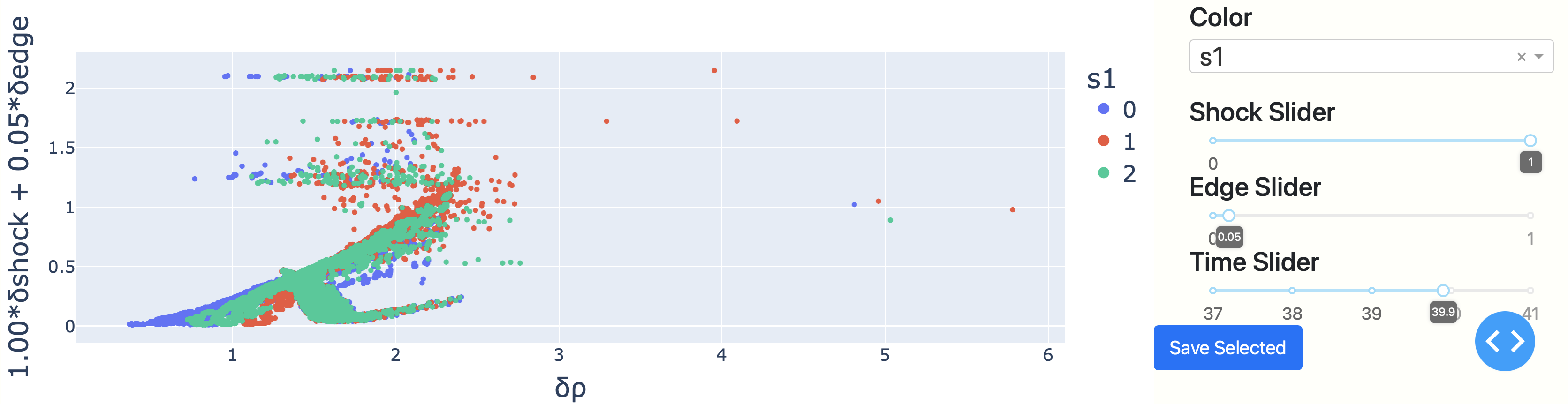}
    \caption{Demonstration of the slider options. In this image, the user has altered the weight of $\delta$edge using a slider and changed the coloring of the scatter plot from `color by profile' to `color by s1'. The user can also change the time step at which the comparison to the ground truth was computed using the Time Slider.}
    \label{fig-selector-slider}
    %\alt{A screenshot of the initial page of the visualization tool. It is the same as the previous figure, except the slider labelled 'Edge Slider' has been moved from 1.0 to 0.05. The scatter plot and y-axis label are modified to reflect this.}
\end{figure}
\begin{figure}[h]
\centering
    \includegraphics[width=0.85\textwidth]{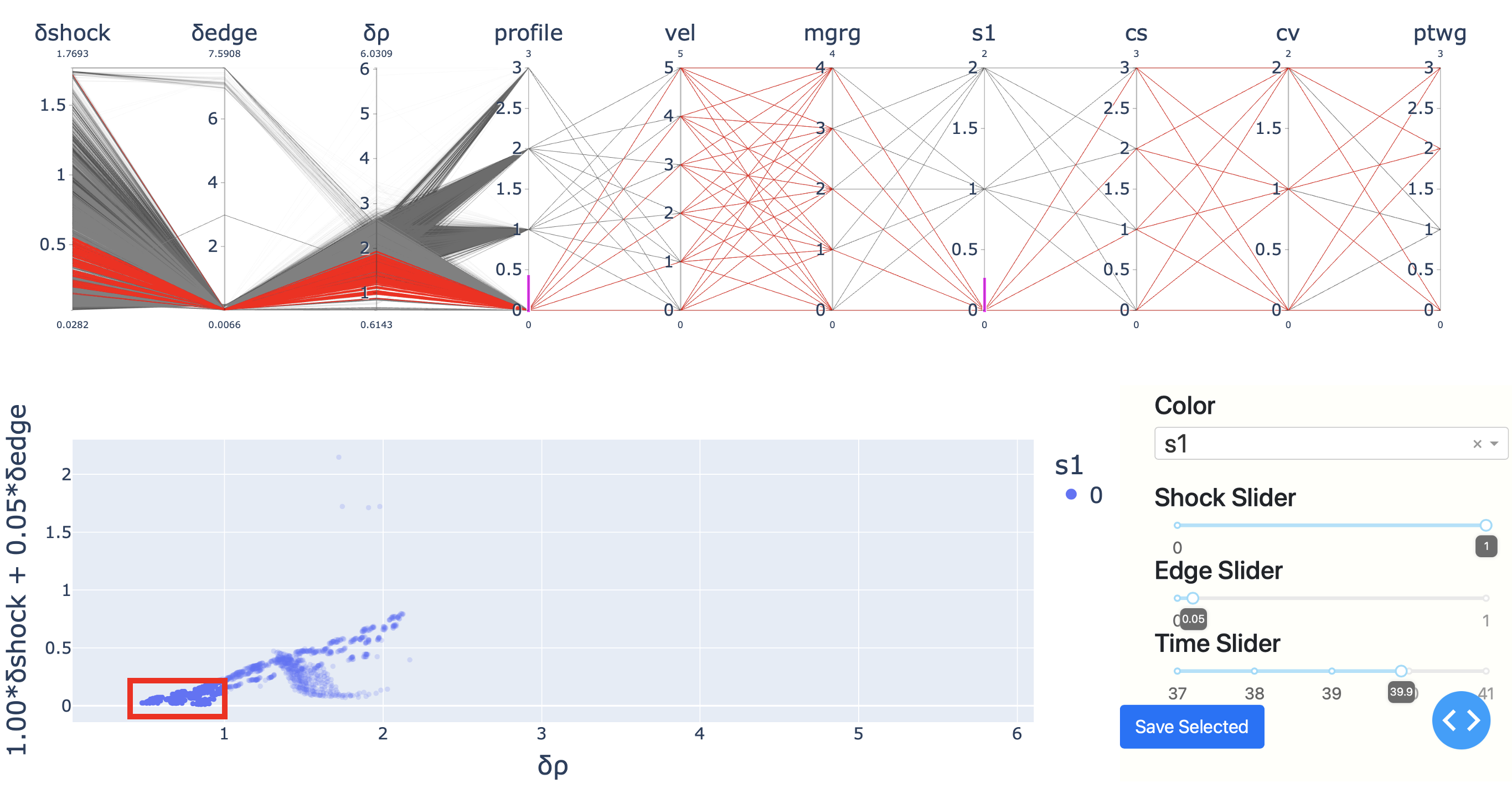}
    \caption{Demonstration of the parameter selection options. The user has selected the ranges profile=0 and s1=0 on the parallel coordinates plot, which filtered the points on the scatter plot. The user has also selected a subset of points on the scatter plot, which highlighted the corresponding lines on the parallel coordinates plot. }
    \label{fig-selector-filter}
    %\alt{A screenshot of the initial page of the visualization tool. It demonstrates the color dropdown, the selection on the scatter plot, and the filter on the parallel coordinates plot. Several points on the scatter plot are selected with a box. On the parallel coordinates plot, a subset of the lines are red, while the rest are gray.}
\end{figure}

\label{sec-gui-description}
The GUI is a web browser-based tool designed to allow the user to 1) visualize the metadata alongside data metrics to aid in understanding of the relationships, 2) select training datasets using visual queries, and 3) visualize the training datasets. The tool is built in Python using the Plotly Dash~\cite{shammamah_hossain-proc-scipy-2019} library.

Figure \ref{fig-selector-screen} shows a screenshot of the first page in the GUI. The header includes a drop-down menu which allows the user to select from one of the pre-computed post-processing methods. Selecting a new method retrieves the data and redraws both graphs. The page includes a parallel coordinates plot, a graph which can concurrently visualize many parameters and support visual queries (subselections) of the data. Each line in the plot represents a single simulation, and the vertical dimensions describe the initial simulation parameters and the comparison against the ground truth. 

At the bottom of the page is a scatter plot of the combination of shock and edge features against the density. The plot can be customized via a dropdown which changes the point color to visualize one of the initial conditions, and there are two sliders which control the weighting of $\delta$shock and $\delta$edge, which are described in Section \ref{sec:usecase}. A third slider allows the user to view the same plot at a different time step; this was added after the users expressed the need to track the evolution of the plot across the time dimension. 

By selecting a subset of the data using the box or lasso selection tool, the user can view the metadata related to the selection in the parallel coordinates plot. Additionally, the user can apply filters to the parallel coordinates plot which filter the scatter plot. The two filtering modes enable comparison within specific initial conditions. These capabilities are demonstrated in Figure ~\ref{fig-selector-filter}.

Lastly, the selected data can be saved as a training dataset. The tool automatically adds the new dataset to the back-end database. It also saves metadata about the selection process such as Date/Time, the type of selection (box versus lasso) and the boundaries of the selection window. It also saves a string describing the filter applied to the parallel coordinates plot; as an example, in Figure \ref{fig-selector-filter}, the user has used the parallel coordinates plot to select simulations with initial conditions `profile $= 0$' and `$s1 = 0$', and the user has selected a box in the bottom left corner of the scatter plot. If the selection is saved using the prompt, the tool will save the metadata filter string ``profile 0; s1 0" alongside the other GUI settings. 

In addition, the save prompt requires a description string and  allows the data to be subsampled with a fixed probability. Requiring a description string forces the user to consider the goal and expected outcomes of training on the new dataset. Additionally, it provides a way to record qualitative decisions, which are a crucial part of the research cycle. The list of training datasets, descriptions, and saved parameters can be viewed as a table with selectable rows in a separate data management page. By selecting a dataset on the table, the user can delete it from the database, save it to a JSON file for further processing, or visualize the points as a table or a graph. 

After a training dataset is saved, the source of the data that was displayed (including the ground truth, computation method, and time step) and the visualization settings (including scatter plot color, shock/edge slider values, and any applied filters) are also saved to the back-end database. The settings of any previously saved training dataset can be retrieved so that the user can reproduce and tweak previous selections. This allows the user to easily resume a previous workflow and make small modifications in order to explore new training datasets based on previous results.

\section{Results}
After using the completed tool, the scientists evaluated how the tool affected their workflow. They specifically noted its usefulness for data exploration. According to the scientist evaluating the tool, ``The tool helps not only to select the relevant simulations, but also to quickly and efficiently visualize individual parameter trends, distinguishing which parameters perturb the features the least, or determining separate clusters of simulations corresponding to multi-modal degeneracy in the data."

\begin{figure}
\centering
\begin{tabular}{cc}
  \includegraphics[width=0.45\textwidth]{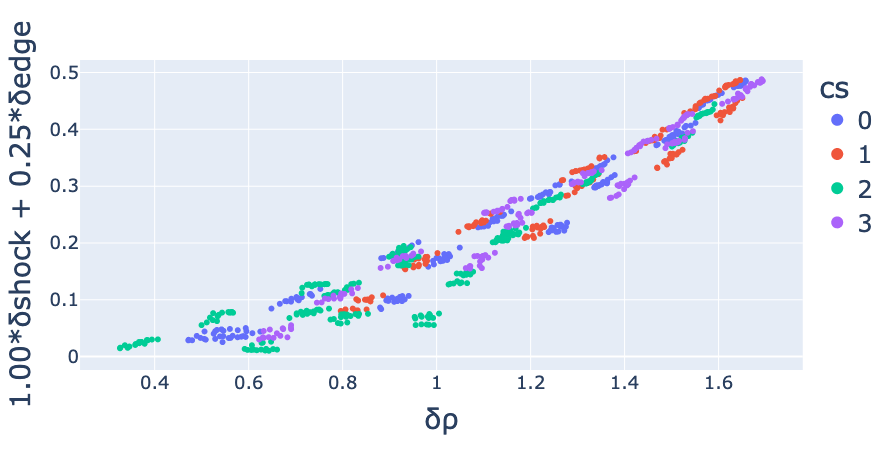} &
  \includegraphics[width=0.45\textwidth]{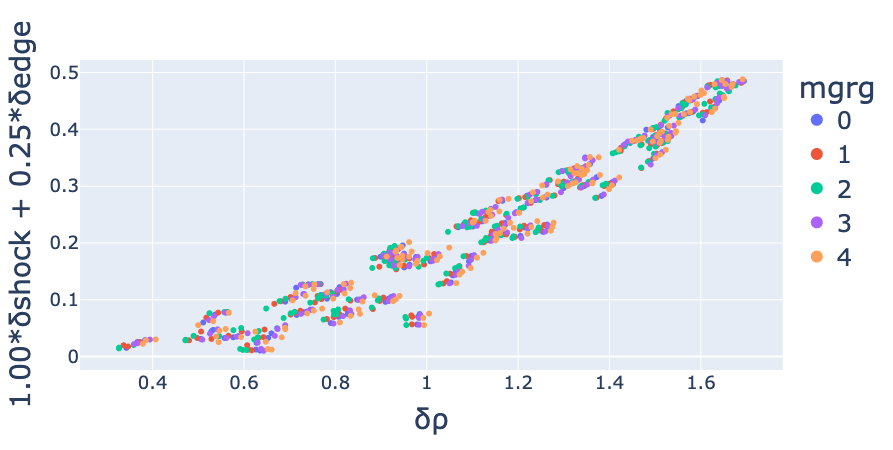}
  %\\
  %\includegraphics[width=0.45\textwidth]{strong_sensitivity_s1.png} &
  %\includegraphics[width=0.45\textwidth]{weak_sensitivity_cv.png}
\end{tabular}
\caption{An example of using the database tool to explore sensitivity to model parameters.
%As in the plots above, individual points correspond to individual simulations in the 
%$\delta\rho$-$\delta$~shock plane (in the $L_2$ norm). 
Color-coded are different discrete values of the parameters: {\tt cs} (left), 
 and {\tt mgrg} (right). The visualization tool
clearly reveals the parameters that strongly impact the feature behavior (left panels),
and those that impact only marginally (right panels). 
Machine learning decisions can be made accordingly.}
\label{fig-parameter-sensitivity}
\end{figure}

We will highlight three key conclusions made by the domain scientists upon exploring the data with this tool. 
\paragraph{Parameter Sensitivity. }The tool allowed the users to easily demonstrate that there is a strong sensitivity to some parameters (e.g. {\tt cs}) which exhibit clear patterns on the figure and therefore machine learning can be applied (see Fig.~\ref{fig-parameter-sensitivity}, left). 
At the same time, sensitivity to other parameters (e.g. {\tt mgrg}) is very weak. Figure~\ref{fig-parameter-sensitivity} (right panel) gives an example of this, demonstrating that simulations marked with different values of the parameter {\tt mgrg} do not form clear trends or separable sets.
It would therefore be meaningless to attempt to recover these parameters from features using any machine learning techniques.
Such parameters are unrecoverable.
\paragraph{Relative effects of predictor variables. }The ability to independently vary the weighting coefficients in the shock and edge differences allowed the users to conclude that the edge difference is too miniscule and noisy to strongly affect the behavior of the scatter plot. Examining different norms showed that the maximal norm is rather noisy, while the $L_1$ norm provides less distinction and separability in different parameter values.
Therefore, it appears that for this datatset using the shock difference and the $L_2$ norm is the best combination for this problem.

\begin{figure}
\centering
\includegraphics[width=\textwidth]{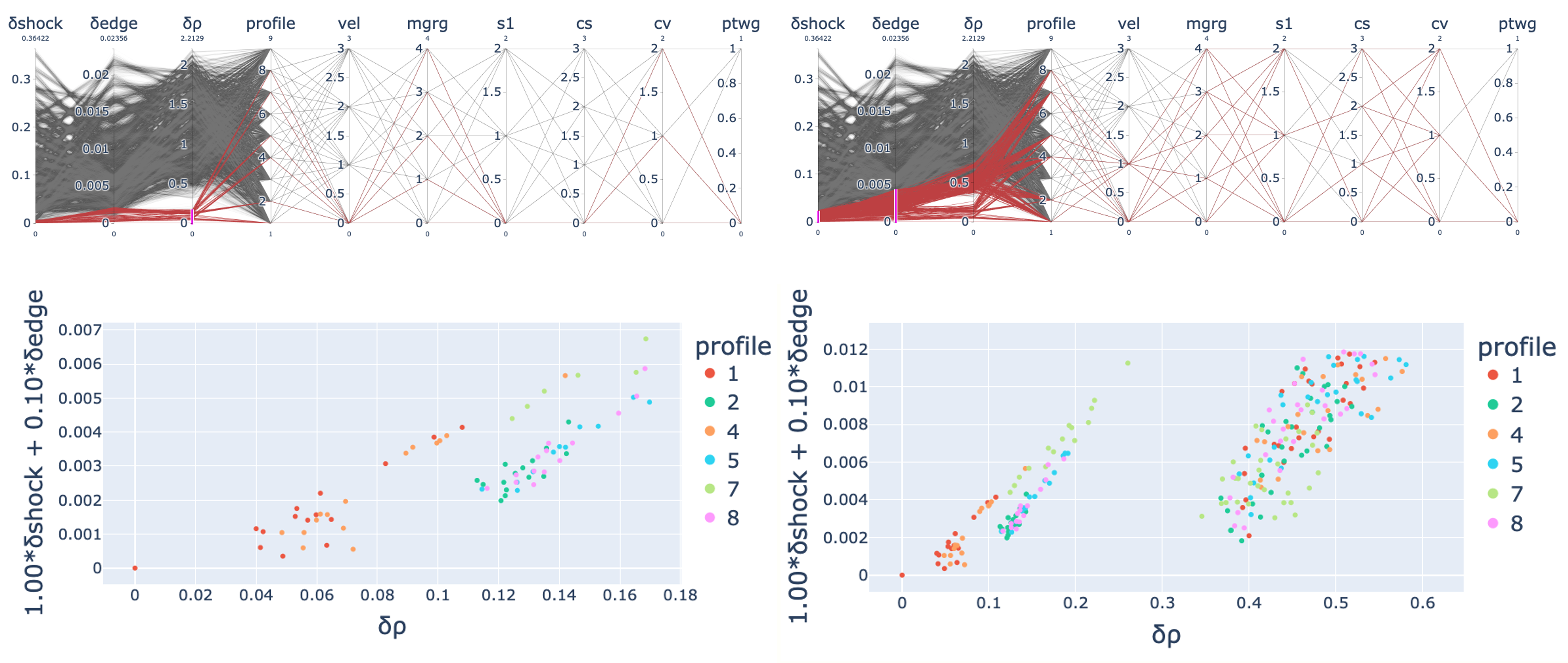}

\caption{Using the parallel coordinates tool to restrict the distance to the ground truth simulation
in density space (left) and in feature space (right). This visualization shows that, a small distance in density space implies a small distance in features space; however,
the opposite is not true: similarity in feature space does not imply that two simulations will 
have similar density profiles.
}
\label{fig-degeneracy-illustration}
\end{figure}

\paragraph{Recoverability of density field. }The central hypothesis of this application project, i.e. whether the density field can be recovered from features, was further explored in detail for different time steps. 
For each time step, scatter plots were visualized with the imposed selection of small distances to the 
ground truth in density space and in feature space (see Fig.~\ref{fig-degeneracy-illustration} for one 
of the selected time step visualizations, $t = 40~\mu$s).
The left panel shows the resulting plot for restricted distance in density space, and the right panel shows 
the same for restricted distance in feature space.
The visualized pattern illustrates that simulations that have a small distance in density space, also have a small distance in feature space. The opposite, however, is not true: a small distance in feature space does not imply that the two  simulations will have similar density distributions.

This illustrates the degeneracy of the problem of density reconstruction from features alone. The visualization tool also suggests a solution to this. The scatter plot, which is color-coded according to the initial perturbation profile, exhibits clear clustering. This demonstrates that the degeneracy of the problem could be partially lifted if this parameter for
the ground truth simulation is known.

\begin{figure}
\centering
\includegraphics[width=0.75\textwidth]{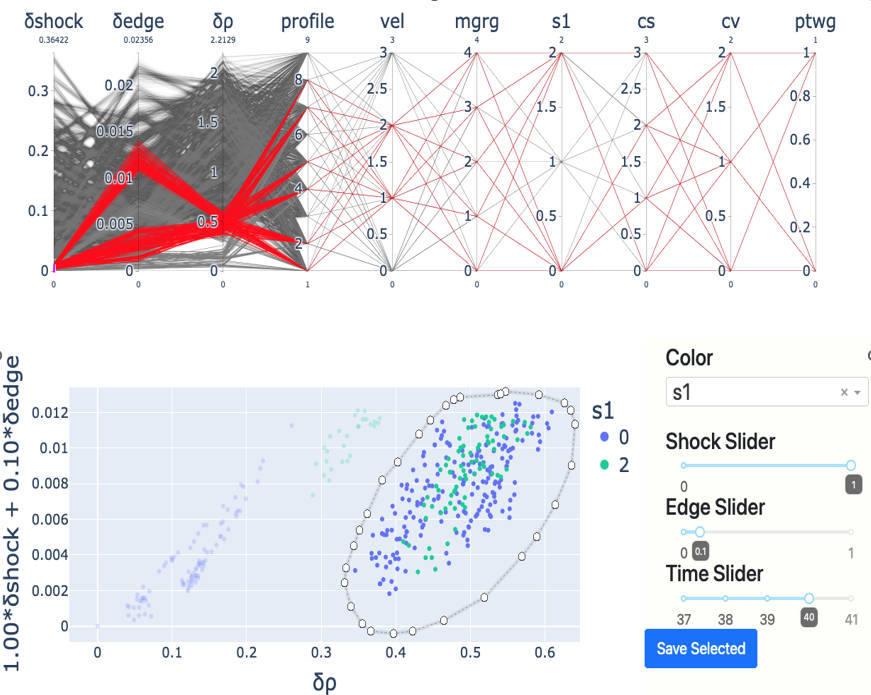}
\caption{An example of a bimodal distribution of simulations with small feature differences. The user selected simulations with small values of $\delta$shock using the parallel coordinates plot. This yields simulations with values of $\delta$edge clustered around two distinct modes.The lasso tool is then used to select only the points
of interest from the scatter plot.}
\label{fig-bimodal-distribution}
\end{figure}

Once these conclusions were made, the tool also allowed the scientists to select relevant subsets of the data for further analysis. To capture the variability in density behavior that corresponds to this degeneracy, the users needed to select
subsets of the local training data. An example of applying the tool to make such selections is
illustrated in Fig.~\ref{fig-bimodal-distribution}.
In this case, for a restricted set of parameters, a subset of simulations with the smallest 
feature distance is selected.
Two clusters of points appear: one with smaller density distance (selected by a rectangular tool),
and another with higher density distance (selected with a lasso tool).
Color-coding the data with different parameter values reveals no clear distinction between the two 
clusters.  This selection is saved to obtain the list of simulations with
these peculiar properties.
Further exploration of individual simulation differences, as well as using them as input to a GAN will
capture the variability in the density profiles and obtain a probabilistic range of density distributions
from the features.

Domain scientists noted that the dynamic visualization tool would allow them to make their analysis far more sophisticated; they reported ``[In previous work] we could not investigate multi-modal regimes of parameter estimation, i.e. when multiple regions in parameter space satisfy the same set of features. Using the proposed metadata tool, it is possible to refine this study and advance it to 2D and 3D."

\section{Discussion}
We have demonstrated a database back-end and visualization tool to support data exploration and training data selection for learning on radiograph data. The tool contributed several concrete improvements over the initial workflow. First, the visualization dashboard enabled interactive exploration of the data, which allowed the scientists to explore hypotheses about the parameter space. Second, the selection interface streamlined the process of selecting training datasets based on visual queries. And thirdly, the integration of a back-end database tracked the necessary metadata parameters to understand and reproduce the training data selection criteria. 

We found that the advances in ML metadata tracking technology in industry have not yet extended to the scientific research cycle. There is a significant gap between the state-of-the-art and the workflow that is implemented in practice by scientists. Our tool, which is geared toward data exploration and the tracking of training data sets, is a first step toward bridging this gap. 

Future work could extend the scope of the tool to encompass the entire machine learning pipeline. The tool described here is intended to support the data exploration and training data selection phase. However, an important metric for a training dataset is the accuracy of the machine learning model trained on it. Future improvements could visualize the model parameters and test accuracy to allow the scientists to look for patterns in training data set efficacy. Additionally, the GUI could connect to the model training and provide support for modifying the training parameters; this would minimize wasted time and human error in transferring data between systems. 

Another key future direction is to modify the tool to support more general scientific experiments. The functionality of the tool was tuned to the needs of the radiograph simulation project highlighted in this paper; however, the problem the tool solves - determining how a response variable can be recovered from several predictor variables - is general. Additionally, we suspect that the functionality, such as local training data selection and cross filtering via parallel coordinates, could prove useful for a variety of projects. 
\section{Acknowledgements}
This work was supported by the U.S. Department of Energy through the Los Alamos National Laboratory (LANL) and the Laboratory Directed Research and Development program of LANL. Los Alamos National Laboratory is operated by Triad National Security, LLC, for the National Nuclear Security Administration of U.S. Department of Energy (Contract No. 89233218CNA000001). The Data Science Infrastructure (DSI) project is part of the Advanced Simulation and Computing (ASC) program.
We would like to thank Soumi De for her feedback in the design of the tool, Marc Klasky for providing access to the simulations database, as well as valuable input, and Terece Turton, Divya Banesh and James Ahrens for their comments on the manuscript.
\printbibliography
\newpage
\appendix
\section{Data Post-Processing}
\label{sec-pipeline}
 The main characteristic aspect of this project is that the features are not fully descriptive of the density field; that is, two simulations that have similar shock and edge locations at a given time step can have large differences in the density space and can have different initial parameters. A technique 
 to incorporate this aspect into the model
 is local training data selection; the model is trained on a dataset consisting of simulations which have large distances in density but small differences in shock and edge. In this section, we outline the data post-processing workflow 
 for computing the density distances using various metrics.

To accomplish this, each simulation is compared against a fixed `Ground Truth' simulation at a fixed time step. Each simulation in the database consists of a density profile, shock and edge at 40 time steps; for each simulation and at each time step, we compute the distance to the ground truth in density, edge location, and shock location. This distance can be computed using any standard norm function, such as the $L_1$, $L_2$, or maximal ($L_\infty$) norms. The following are the equations for evaluating these norms on a uniform 2D cylindrical grid $\{R, z\}$ with spacing $\{\Delta R, \Delta z\}$ in the radial and cylindrical direction. Let $\rho_{gt}$ be the ground truth density profile, and $\rho$ be the simulation being compared:
\begin{align}
 \bar{D}_1(\rho_{gt},\rho) 
 &= \frac{2\pi \Delta R\ \Delta z}{V^+}\sum_{j,k\in\mathcal{D}^+} 
 \left|\rho_{gt}(R_j, z_k) - \rho(R_j, z_k)\right| R_j,
 \\
 \bar{D}_2(\rho_{gt},\rho) 
 &= \left(\frac{2\pi \Delta R\ \Delta z}{V^+}\sum_{j,k\in\mathcal{D}^+} \left[\rho_{gt}(R_j, z_k) - \rho(R_j, z_k)\right]^2 R_j\right)^{1/2},
 \\
 \bar{D}_\infty(\rho_{gt},\rho) 
 &= \max_{j,k\in\mathcal{D}^+} \left|\rho_{gt}(R_j, z_k) - \rho(R_j, z_k)\right|.
\end{align}
Here, the domain $\mathcal{D}^+$ is defined as an intersection of points where
both densities are nonzero: $\rho_{gt}\rho>0$, and $V^+$ is the 3D volume of this domain.

The distance to the ground truth shock and edge can be computed in an analogous way; the shock and edge are represented as the coefficients of a Fourier decomposition, and the distance in the coefficient space is computed. 

Therefore, the post-processing pipeline produces a single floating-point number for each time step representing the distance in shock, edge, and density to the ground truth. However, the distance in density, shock and edge may be computed using several different methods for each simulation. 

These distances can be visualized and used to make training data selections. The selection step may be iterated depending on feedback from later points in the pipeline.

\section{Database Structure} \label{sec:db}
\begin{figure}[h!]
\centering
\includegraphics[width=\textwidth]{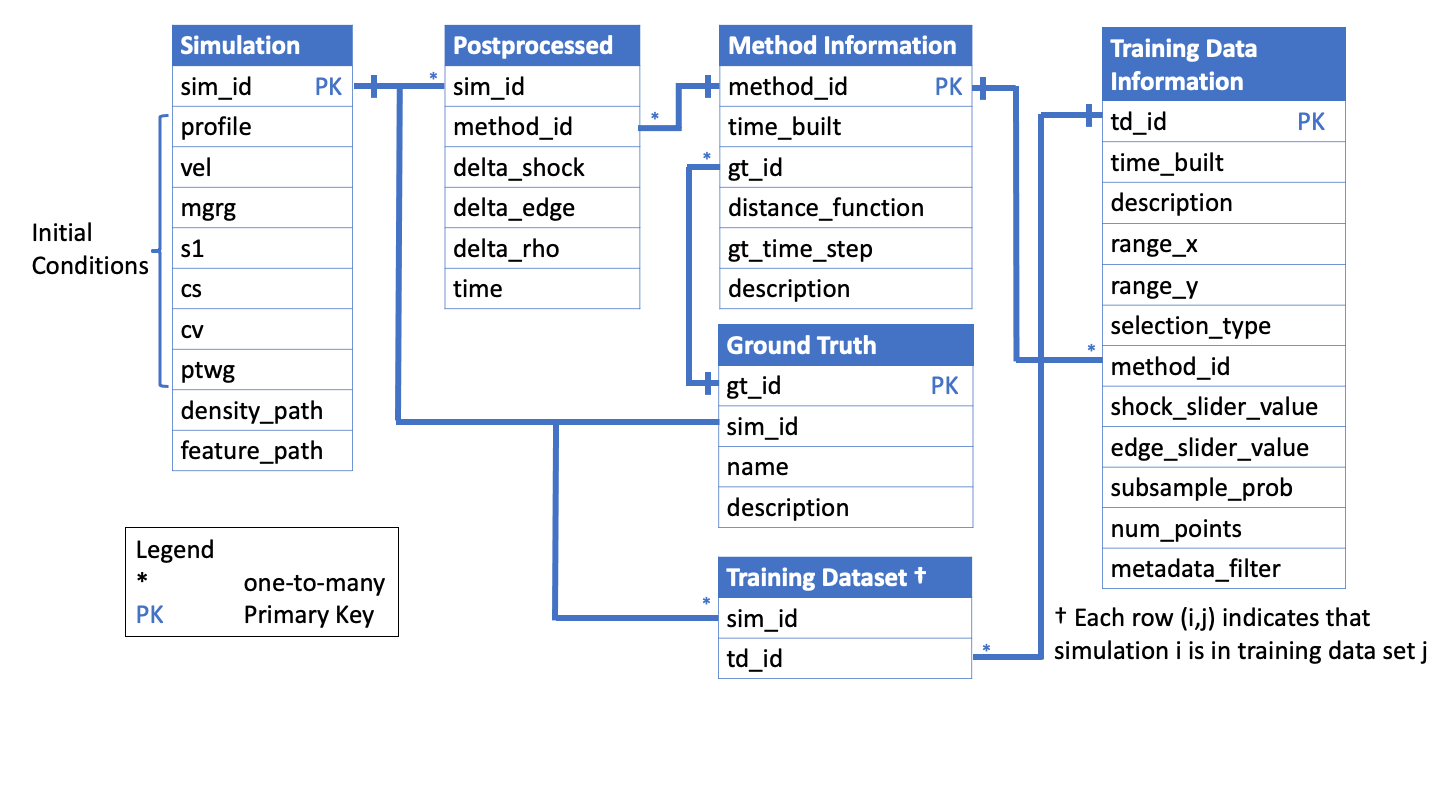}
\vspace*{-18mm} 
\caption{A database structure diagram illustrating the organization of the database. The boxes represent tables in the database, and each listed name is a field in the table. PK represents the primary key, a unique identifier allowing the same simulations to be linked across tables. When a * is present, the same ID may appear as a row several times in the table. The database houses hundreds of thousands of rows.}
\label{fig-database-structure}
\end{figure}

In this section, we introduce a database structure pictured in Figure~\ref{fig-database-structure} which houses all the metadata related to the simulations, post-processing steps, and decisions made in producing training data. The database is built and accessed through SQLite~\cite{sqlite}. In combination with the visualization tool, this database contains all the information needed to reproduce the training data from the original simulation repository. Additionally, whenever a qualitative decision is made, the database also houses a description string that justifies the decision. 

The `Simulation' table contains metadata about the simulation data itself. Each simulation is generated from a set of seven initial conditions, which are grouped into categorical values. Each row contains a unique ID, the value of each of the initial conditions used to generate it, and a pointer to the locations of the density and feature files. 

The Post-processed Data table contains the metrics that are computed from the data using the method described in Appendix~\ref{sec-pipeline}. These metrics are necessary to choose the training dataset, but they take significant time to generate (and thus cannot be generated on the fly by the visualization tool). Therefore, we store the post-processed data in the database so it can be easily retrieved and visualized. 

The Method Information table contains the metadata describing how the post-processed data was computed. In order to generate the post-processed data, the user must select a ground truth simulation, a time step to compare against, and the norm used to compute the distance. Each time the data is post-processed, these parameters are stored in the Method Information table and linked to the Post-processed data table using a unique ID. The database supports the ability to choose multiple ground truths, and to compare against a given ground truth using multiple time steps and norms; therefore, the simulation used as the ground truth is stored in a separate table and assigned a unique `Ground Truth ID'. This is linked to the information table on the computation method. Lastly, the table has the option to store a description string about each method. 

The Training Dataset table contains the IDs of the simulations which are included in training. Again, each training dataset receives its own ID, and metadata is stored in an information table. The training data is selected via a visual query, so the query parameters are stored alongside a description string. Section \ref{sec-gui-description} describes the visual query capability in more detail.
\end{document}